\documentclass[10pt,twocolumn,letterpaper]{article}

\usepackage[pagenumbers]{cvpr} 

\usepackage{graphicx}
\usepackage{amsmath}
\usepackage{amssymb}
\usepackage{booktabs}
\usepackage{makecell}
\usepackage{float}
\usepackage{color}

\usepackage[pagebackref,breaklinks,colorlinks]{hyperref}

\usepackage[capitalize]{cleveref}
\crefname{section}{Sec.}{Secs.}
\Crefname{section}{Section}{Sections}
\Crefname{table}{Table}{Tables}
\crefname{table}{Tab.}{Tabs.}

\def\cvprPaperID{3942} 
\def\confName{CVPR}
\def\confYear{2022}

\begin{document}

\title{Deep Rectangling for Image Stitching: A Learning Baseline}


\author{Lang Nie$^{1,2}$, Chunyu Lin$^{1,2}$\thanks{Corresponding author}, Kang Liao$^{1,2}$, Shuaicheng Liu$^{3}$, Yao Zhao$^{1,2}$\\
$^{1}$Institute of Information Science, Beijing Jiaotong University, Beijing, China\\
$^{2}$Beijing Key Laboratory of Advanced Information Science and Network, Beijing, China\\
$^{3}$University of Electronic Science and Technology of China, Chengdu, China\\
\url{https://github.com/nie-lang/DeepRectangling}
}


\twocolumn[{%
\renewcommand\twocolumn[1][]{#1}%
\maketitle
\begin{center}
    \centering
    \vspace{-0.5cm}
    \captionsetup{type=figure}
    \includegraphics[width=.95\textwidth]{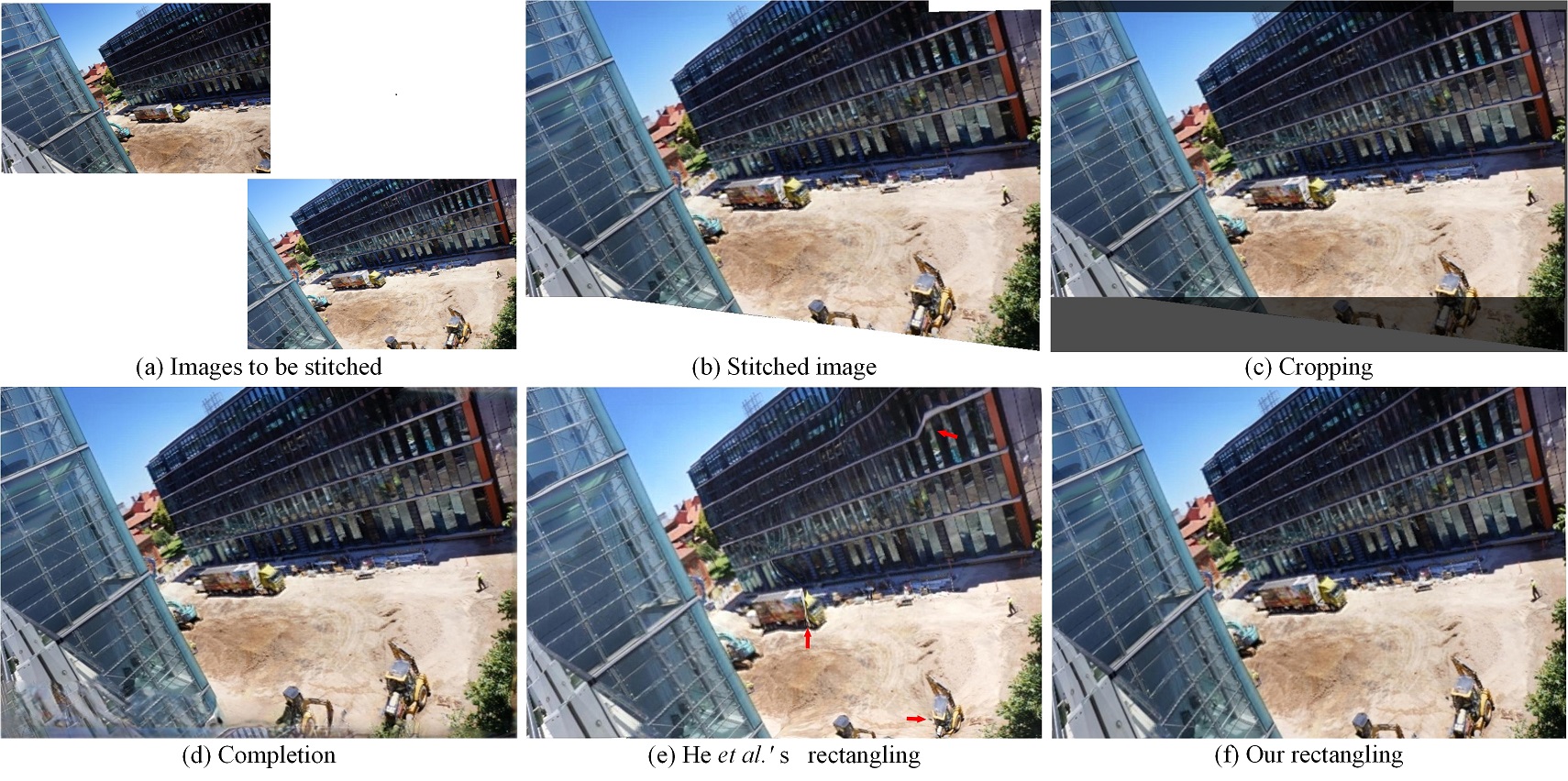}
    \vspace{-0.3cm}
    \captionof{figure}{Different solutions to irregular boundaries in image stitching. (a) A classic image stitching dataset that is not included in the proposed dataset (APAP-conssite\cite{zaragoza2013projective}). (b) Stitching images using UDIS\cite{nie2021unsupervised} with inevitable irregular boundaries. (c) Cropping the boundaries to get a rectangular image. (d) Completing the missing regions using LaMa\cite{suvorov2021resolution}. (e) He $et\ al.$'s rectangling \cite{he2013rectangling} with noticeable distortions. (f) Our rectangling with high content fidelity.}
    \label{fig1}
\end{center}%
}]


\begin{abstract}
  Stitched images provide a wide field-of-view (FoV) but suffer from unpleasant irregular boundaries.
  To deal with this problem, existing image rectangling methods devote to searching an initial mesh and optimizing a target mesh to form the mesh deformation in two stages. Then rectangular images can be generated by warping stitched images. However, these solutions only work for images with rich linear structures, leading to noticeable distortions for portraits and landscapes with non-linear objects.

  In this paper, we address these issues by proposing the first deep learning solution to image rectangling. Concretely, we predefine a rigid target mesh and only estimate an initial mesh to form the mesh deformation, contributing to a compact one-stage solution.
  The initial mesh is predicted using a fully convolutional network with a residual progressive regression strategy.
  To obtain results with high content fidelity, a comprehensive objective function is proposed to simultaneously encourage the boundary rectangular, mesh shape-preserving, and content perceptually natural.
  Besides, we build the first image stitching rectangling dataset with a large diversity in irregular boundaries and scenes.
  Experiments demonstrate our superiority over traditional methods both quantitatively and qualitatively. 

\end{abstract}

\section{Introduction}
\label{sec:introduction}
Image stitching algorithm \cite{zaragoza2013projective, lee2020warping, lin2011smoothly, chen2016natural} can generate a wide FoV image (Fig.\ref{fig1}{\color{red}b}) from normal FoV images (Fig.\ref{fig1}{\color{red}a}). These methods optimize a global or local warp to align the overlapping regions of different images. Nevertheless, non-overlapping regions always suffer from irregular boundaries \cite{chang2014shape}. People who use image stitching technology have to be tolerant of unpleasant boundaries.

To deal with the irregular boundaries, one of the solutions is to crop a stitched image with a rectangle. However, cropping inevitably reduces the FoV of the stitched image, which contradicts the original intention of image stitching. Fig.\ref{fig1}{\color{red}c} demonstrates an example, where the dark regions indicate the discarded areas by cropping. On the other hand, image completion can synthesize the missing regions to form a rectangular image. Nevertheless, there is currently no work to design a mask for irregular boundaries in image stitching, and even SOTA completion works \cite{suvorov2021resolution, teterwak2019boundless} show unsatisfying performance (Fig.\ref{fig1}{\color{red}d}) when processing the stitched images. Moreover, the completion methods may add some contents that seem to be harmonious but different from reality, making them unreliable in high-security applications such as autonomous driving \cite{lai2019video}.

To obtain a rectangular image with high content fidelity, image rectangling methods \cite{he2013rectangling, 7298617, he2013content} are proposed to warp a stitched image to a rectangle via mesh deformation. However, these solutions can only preserve structures with straight/geodesic lines such as buildings, boxes, pillars, etc.
For non-linear structures such as portraits \cite{9578096}, distortions are usually generated.
Actually, the capability to preserve linear structures is limited by line detection, thus distortions also occur in linear structures sometimes (Fig.\ref{fig1}{\color{red}e}).
Moreover, these traditional methods are two-stage solutions that search an initial mesh and optimize a target mesh successively, making it challenging to be parallelly accelerated.

To address the above problems, we propose the first one-stage learning baseline, in which we predefine a rigid target mesh and only predict an initial mesh.
Specifically, we design a simple but effective fully convolutional network to estimate a content-aware initial mesh from a stitched image with a residual progressive regression strategy.
Besides, a comprehensive objective function consisting of a boundary term, a mesh term, and a content term is proposed to simultaneously encourage the boundary rectangular, mesh shape-preserving, and content perceptually natural.
Compared with the existing methods, our content-preserving capability is more general (not limited to linear structures) and more robust (Fig.\ref{fig1}{\color{red}f}) due to the effective semantic perception in our content constraint.

As there is no proper dataset readily available, we build a deep image rectangling dataset (DIR-D) to supervise our training.
First, we apply He $et\ al.$'s rectangling \cite{he2013rectangling} to real stitched images to generate synthetic rectangular images. Then we utilize the inverse of rectangling transformations to warp real rectangular images to synthetic stitched images.
Finally, we manually filter out images without distortions from tens of thousands synthetic images for several epochs strictly, yielding a dataset with 6,358 samples.

Experimental results show that our approach can generate content-preserving rectangular images efficiently and effectively, outperforming the existing solutions both quantitatively and qualitatively.
To sum up, we conclude our contributions as follows:
\begin{itemize}
    \item We propose the first deep rectangling solution for image stitching, which can effectively generate rectangular images in a residual progressive manner.
    \item Existing methods are two-stage solutions while ours is a one-stage solution, enabling efficient parallel computation with a predefined rigid target mesh. Besides, ours can preserve both linear and non-linear structures.
    \item As there is no proper dataset of pairs of stitched images and rectangular images, we build a deep image rectangling dataset with a wide range of irregular boundaries and scenes.
 \end{itemize}

\section{Related Work}
\label{sec:related_work}
This paper offers a deep learning based rectangling solution for image stitching. Hence, this section reviews previous works related to image stitching and image rectangling.
\subsection{Image Stitching}
\label{subsec:image_stitching}
\vspace{-2pt}
Aligning overlapping regions \cite{li2019local} is the core goal of image stitching. But to produce natural stitched images, it is also necessary to minimize projective distortions of non-overlapping regions.
In \cite{chang2014shape, lin2015adaptive}, the projective transformation of overlapping regions is smoothly extrapolated into non-overlapping regions, and the resultant warp gradually changes from projection to similarity across the image. Li $et\ al.$ \cite{li2017quasi} propose a quasi-homography warp, which relies on a global homography while squeezing non-overlapping areas. Liao and Li \cite{liao2019single} propose two single-perspective warps to preserve perspective consistency with reduced projective distortions. Recently, Jia $et\ al.$ \cite{jia2021leveraging} consider the scenes of long lines and keep the shape of global co-linear line segments during the stitching process.

\vspace{1pt}

Although the existing image stitching algorithms can reduce projective distortions and keep the natural appearance, they can not solve the problem of irregular boundaries in stitched images.

\begin{figure*}
  \centering
  \begin{subfigure}{0.925\textwidth}
    \centering
    \includegraphics[width=0.97\textwidth,height=2.2cm]{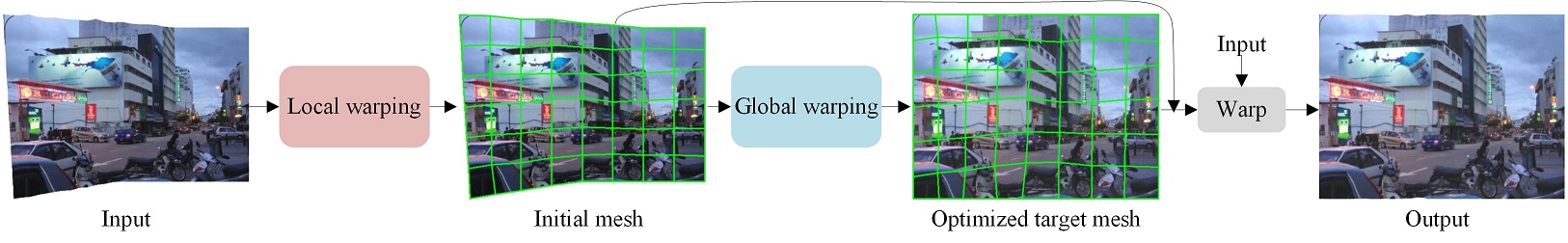}
    \caption{Two-stage traditional baseline (only 8$\times$6 mesh is drawn for clarity).}
    \label{fig2-a}
  \end{subfigure}
  \hfill
  \begin{subfigure}{0.8\textwidth}
    \includegraphics[width=0.97\textwidth,height=2.2cm]{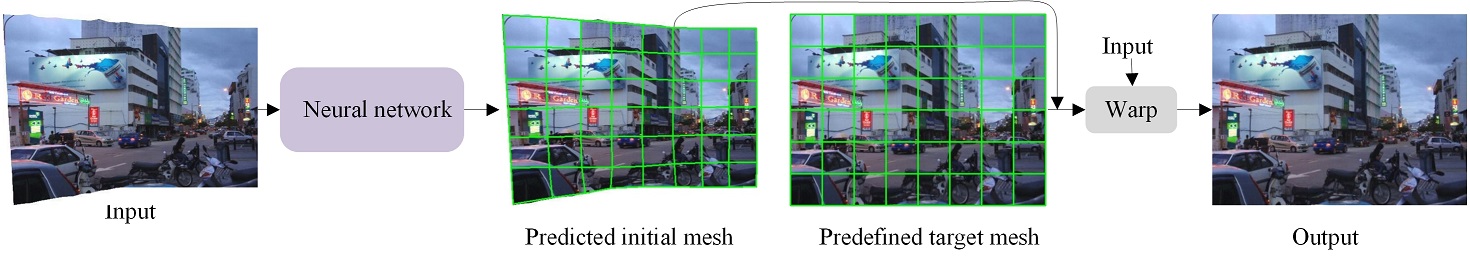}
    \caption{The proposed one-stage learning baseline.}
    \label{fig2-b}
  \end{subfigure}
  \vspace{-0.2cm}
  \caption{Traditional baseline vs. learning baseline. Traditional baseline solves the rectangling warp in two stages by searching the initial mesh and optimizing the target mesh successively, while our solution solves it in one stage because ours only predicts the initial mesh.}
  \label{fig2}
  \vspace{-0.3cm}
\end{figure*}

\subsection{Image Rectangling}
\label{subsec:image_Rectangling}
To get rectangular stitched images, He $et\ al.$ \cite{he2013rectangling} propose to optimize a line-preserving mesh deformation. However, the proposed energy function can only preserve linear structures. Considering straight lines may be bent in a panorama (ERP format), Li $et\ al.$ \cite{7298617} improve the line-preserving energy term into the geodesic-preserving energy term. But this improvement limits its application to the panorama and the geodesic lines can not be detected from a stitched image directly. Later, Zhang $et\ al.$ \cite{zhang2020content} bridge the image rectangling and image stitching in a unified optimization. Nevertheless, to reduce distortions of the final rectangular result, they make a compromise to the rectangular shape, adopting piecewise rectangular boundary constraints instead.

Image rectangling is rarely studied because the unstable performance and heavy time-consumption make it impractical in applications. In this paper, we propose a simple but effective learning baseline to address these issues.

\section{Methodology}
\label{sec:methodology}
We first analyse differences bewtween the traditional baseline and the proposed learning baseline in Section \ref{subsec:baseline_comparison}. Then, our network structure and the objective function are discussed in Section \ref{subsec:network_structure} and Section \ref{subsec:objective_function}, respectively.

\subsection{Traditional Baseline vs. Learning Baseline}
\label{subsec:baseline_comparison}
A rectangling solution should solve the initial mesh and the target mesh to form the mesh deformation. Then the rectangling result can be obtained via warping.

\vspace{-10pt}
\subsubsection{Traditional Baseline}
\vspace{-5pt}
In classic traditional methods \cite{he2013rectangling, 7298617}, two stages are required: local stage and global stage (shown in Fig. \ref{fig2-a}).

\textbf{Stage 1: local stage.} First, insert abundant seams into the stitched image to get a preliminary rectangular image using seam carving algorithm\cite{avidan2007seam}. Then, place a regular mesh on the preliminary rectangular image and remove all the seams to get an initial mesh for a stitched image with irregular boundaries.

\textbf{Stage 2: global stage.} This stage solves the optimal target mesh via optimizing an energy function to preserve limited perceptual properties such as straight lines.

They produce the rectangular image by warping the stitched image from the initial mesh to the target mesh.

\vspace{-10pt}
\subsubsection{Learning Baseline}
\vspace{-5pt}
As shown in Fig. \ref{fig2-b}, the proposed learning baseline is a one-stage solution.

Given a stitched image, our solution only needs to predict a content-aware initial mesh via a neural network. As for the target mesh, we predefine it to have a rigid shape. Moreover, the rigid mesh shape can enable the acceleration of the backward interpolation using the matrix computation easily \cite{9605632}. Rectangular images can be obtained by warping stitched images from the predicted initial mesh to the predefined target mesh.

Compared with the traditional baseline, the learning baseline is more efficient due to the one-stage pipeline. The content-preserving capability makes our rectangling results more natural in perception (explained in Section \ref{subsubsec:content}).

\subsection{Network Structure}
\label{subsec:network_structure}
\begin{figure*}[!t]
  \centering
  \includegraphics[width=0.95\textwidth]{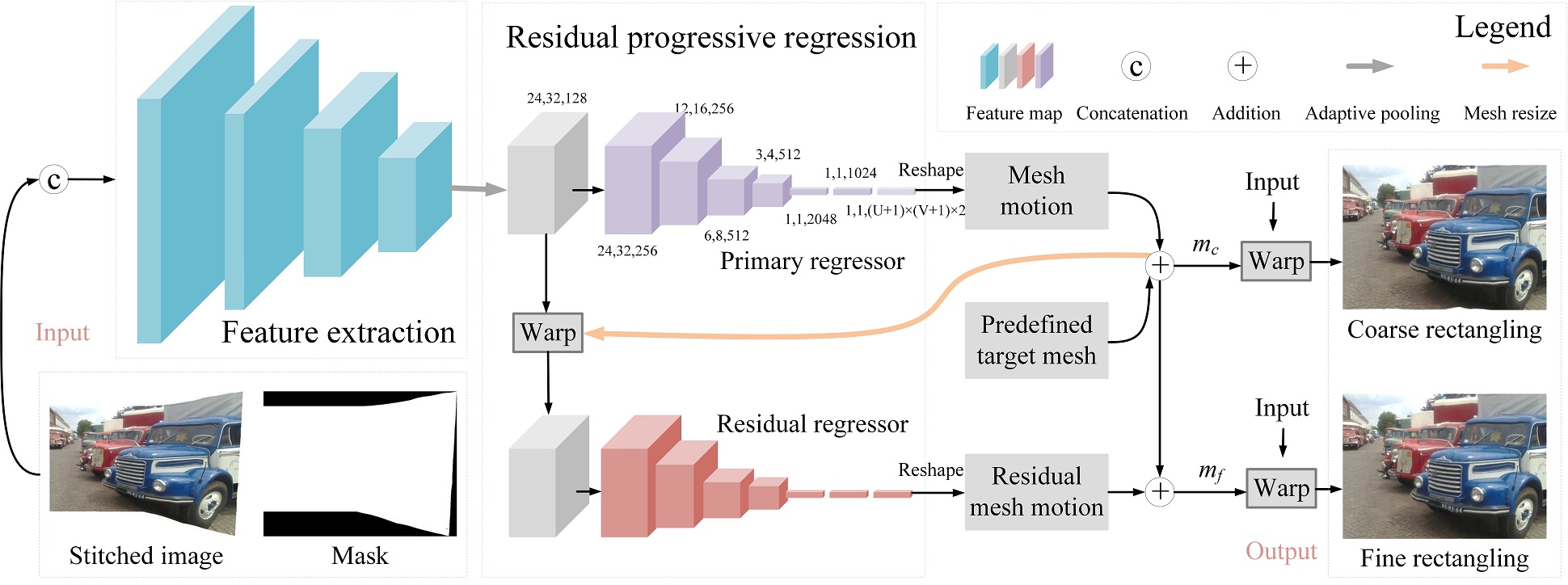}
  \vspace{-8pt}
  \caption{The overall structure of our network which takes a stitched image and a mask as input. It outputs the predicted mesh motions in a progressive manner. The rectangling results can be produced by warping the stitched image with the predicted warps.}
  \label{fig:network}
  \vspace{-0.3cm}
\end{figure*}

Similar to image completion tasks \cite{teterwak2019boundless, suvorov2021resolution}, a stitched mask is also included in the input of the proposed network. As illustrated in Fig. \ref{fig:network}, we concatenate the stitched image $I$ and mask $M$ on the channel dimension as the input. The output is the predicted mesh motions.

\vspace{5pt}
\textbf{Feature extractor.} We stack simple convolution-pooling blocks to extract high-level semantic features from the input. Formally, 8 convolutional layers are adopted, whose filter numbers are set to 64, 64, 64, 64, 128, 128, 128, and 128, respectively. The max-pooling layers are used after the $2$-th, $4$-th, and $6$-th convolutional layers.

\vspace{5pt}
\textbf{Mesh motion regressor.} After feature extraction, an adaptive pooling layer is utilized to fix the resolution of feature maps. Subsequently, we design a fully convolutional structure as the mesh motion regressor to predict the horizontal and vertical motions of every vertex based on the regular mesh.
Supposing the mesh resolution is $U\times V$, the size of the output volume is $(U+1)\times (V+1)\times 2$.

\vspace{5pt}
\textbf{Residual progressive regression.} Observing that the warped result can be regarded as the network input again, we design a residual progressive regression strategy to estimate accurate mesh motions through a progressive manner.
First, we do not use the warped image as the input of a new network directly, because this would double the computational complexity. On the contrary, we warp the intermediate feature maps instead, improving the performance with a slight increase in the computation.
Then, we design two regressors with the same structure to predict primary mesh motions and residual mesh motions, respectively. Although they share the same structure, they are specified for different tasks due to the different input features.

\subsection{Objective Function}
\label{subsec:objective_function}
We optimize our network parameters using a comprehensive objective function that consists of three terms. The optimization goal can be formulated as follows:
\begin{eqnarray}
  L_{total} = \ell_b + \ell_m + \ell_c,
\end{eqnarray}
where $\ell_b$, $\ell_m$, and $\ell_c$ are the boundary term, mesh term, and content term, respectively.

\vspace{-10pt}
\subsubsection{Content Term}
\vspace{-5pt}
\label{subsubsec:content}
The traditional methods \cite{he2013rectangling, 7298617} preserve image contents by preserving the angles of straight/geodesic lines, failing to deal with other non-linear structures. To overcome it, we propose to learn the content-preserving capability from two different perspectives.

\textbf{Appearance loss.} Given the predicted primary mesh $m_p$ and final mesh $m_f$, we enforce the rectangling results to be close to the rectangling labels $R$ in appearance as follows:
\begin{eqnarray}
  \ell_{c}^{a} = \left \| R-\mathcal{W}(I, m_p) \right \|_1 + \left \| R-\mathcal{W}(I, m_f) \right \|_1,
\end{eqnarray}
where $\mathcal{W}(\cdot,\cdot)$ is the warp operation.

\textbf{Perception loss.} To make our results perceptually natural, we minimize the $L2$ distance between rectangling results and labels in high-level semantic perception as Eq. \ref{eq3}:
\begin{eqnarray}
\begin{aligned}
  \ell_{c}^{p} &= \left \| \varphi(R)-\varphi(\mathcal{W}(I, m_c)) \right \|_2 \\
  &+ \left \| \varphi(R)-\varphi(\mathcal{W}(I, m_f)) \right \|_2,
\end{aligned}
\label{eq3}
\end{eqnarray}
where $\varphi(\cdot)$ represent the operation of feature extraction from the '$conv4\_2$' layer of VGG19 \cite{simonyan2014very}. In this manner, various perceptual properties (not limited to linear structures) can be perceived.

\vspace{5pt}
In sum, the content loss is formed by simultaneously emphasizing the similarity in appearance and semantic perception as follows:
\begin{eqnarray}
  \ell_{c} = \omega_a\ell_{c}^{a} + \omega_p\ell_{c}^{p},
\end{eqnarray}
where $\omega_a$ and $\omega_p$ denote the weights for the appearance loss and the perception loss.

\vspace{-10pt}
\subsubsection{Mesh Term}
\vspace{-5pt}
To prevent content distortions in rectangular images, The predicted mesh should not be exaggeratedly deformed. Therefore, we design an intra-grid constraint and an inter-grid constraint to keep the shape of the deformed mesh.

\begin{figure}
  \centering
  \begin{subfigure}{0.16\textwidth}
    \centering
    \includegraphics[width=1\textwidth]{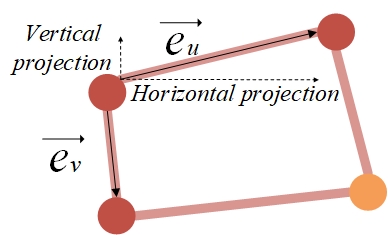}
    \caption{Intra-grid constraint.}
    \label{intra}
  \end{subfigure}
  \hspace{0.8cm}
  \begin{subfigure}{0.16\textwidth}
    \centering
    \includegraphics[width=1\textwidth]{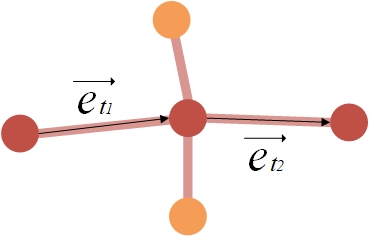}
    \caption{Inter-grid constraint.}
    \label{inter}
  \end{subfigure}
  \vspace{-0.3cm}
  \caption{Mesh shape-preserving constraint.}
  \label{fig4}
  \vspace{-0.3cm}
\end{figure}

\textbf{Intra-grid constraint.} In a grid, we impose constraints on the magnitude and direction of grid edges. As shown in Fig. \ref{intra}, we encourage the direction of the horizontal projection of each horizontal edge $\vec{e}_u$ to the right, together with its norm greater than a threshold $\alpha \frac{W}{V}$ (suppose the stitched image has the resolution of $H\times W$). We use a penalty $P_{hor}$ to describe this constraint as follows:
\begin{eqnarray}
  P_{hor}=\left
  \{\begin{array}{ll}
    \alpha \frac{W}{V}-\langle \vec{e}_u,\vec{i}\rangle, & \langle \vec{e}_u,\vec{i}\rangle  < \alpha \frac{W}{V}  \\
  0, & \langle \vec{e}_u,\vec{i}\rangle  \geq  \alpha \frac{W}{V}
\end{array} \right.
\end{eqnarray}
where $i$ is the horizontal unit vector to the right. As for the vertical edge $\vec{e}_v$ in every grid, we impose a similar penalty $P_{ver}$ as follows:
\begin{eqnarray}
  P_{ver}=\left
  \{\begin{array}{ll}
    \alpha \frac{H}{U}-\langle \vec{e}_v,\vec{j}\rangle, & \langle \vec{e}_u,\vec{i}\rangle  < \alpha \frac{H}{U}  \\
  0, & \langle \vec{e}_v,\vec{j}\rangle  \geq  \alpha \frac{H}{U}
\end{array} \right.
\end{eqnarray}
where $j$ is the vertical unit vector to the bottom. Then, the intra-grid mesh loss is formed using Eq. \ref{eq7}, which can effectively prevent the intra-grid shape from distortions.
\begin{equation}
\begin{aligned}
  \ell_{m}^{intra}&= \frac{1}{(U+1)\times V}\sum_{\vec{e}_u\in m_p\cup m_f}P_{hor} \\ &+ \frac{1}{U\times (V+1)}\sum_{\vec{e}_v\in m_p\cup m_f}P_{ver}.
\end{aligned}
  \label{eq7}
\end{equation}

\textbf{Inter-grid constraint.} We also adopt the inter-grid constraint to encourage neighboring grids to transform consistently. As shown in Fig. \ref{inter}, two successive deformed grid edges $\{\vec{e}_{t1}, \vec{e}_{t2}\}$ are encouraged to be co-linear.
\begin{equation}
  \ell_{m}^{inter}= \frac{1}{N}\sum_{\{\vec{e}_{t1}, \vec{e}_{t2}\}\in m_p\cup m_f}(1-\frac{\langle \vec{e}_{t1},\vec{e}_{t2}\rangle}{\parallel \vec{e}_{t1}\parallel \cdot \parallel \vec{e}_{t2}\parallel }).
\end{equation}
We formulate the inter-grid mesh loss as above, where N is the number of tuples of two successive edges in a mesh.

\vspace{5pt}
In sum, the total mesh term is concluded as follows:
\begin{eqnarray}
  \ell_{m} = \ell_{m}^{intra} + \ell_{m}^{inter},
\end{eqnarray}

\vspace{-10pt}
\subsubsection{Boundary Term}
\vspace{-5pt}
As for the boundary term, we constrain the mask instead of the predicted mesh.
Given a 0-1 mask of a stitched image (as shown in Fig. \ref{fig:network}), we warp the mask and constrain the warped mask close to an all-one matrix $E$ as follows:
\begin{eqnarray}
  \ell_{b} = \left \| E-\mathcal{W}(M, m_p) \right \|_1+ \left \| E-\mathcal{W}(M, m_f) \right \|_1.
\end{eqnarray}

\section{Data Preparation}
\label{sec:data_preparation}
To train a deep image rectangling network, we build an image rectangling dataset (DIR-D), in which each sample is a triplet consisting of a stitched image ($I$), a mask ($M$), and a rectangling label ($R$). We prepare this dataset by the following steps:

\begin{figure}[!t]
  \centering
  \includegraphics[width=0.5\textwidth]{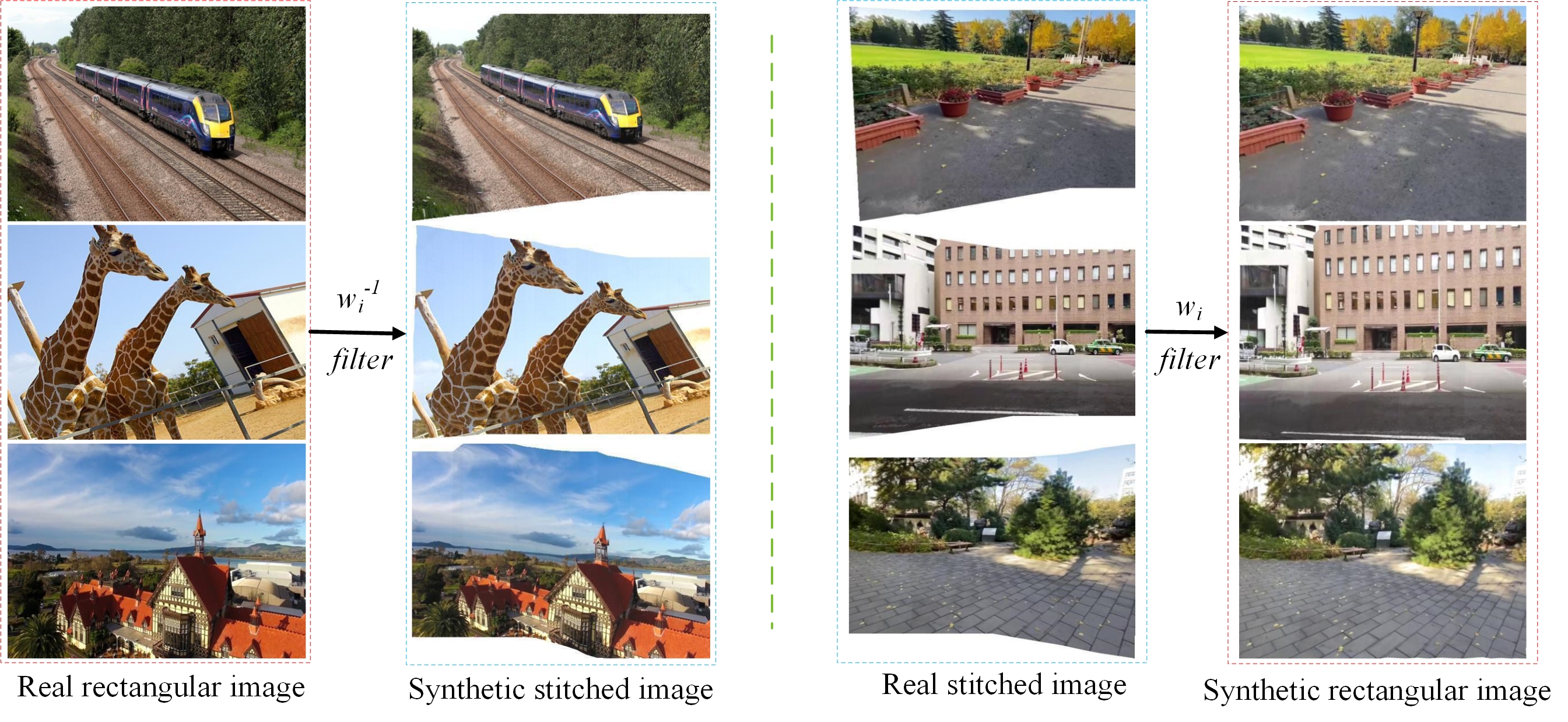} 
  \vspace{-0.7cm}
  \caption{Dataset preparation. Left: synthesize stitched images from real rectangular images. Right: synthesize rectangular images from real stitched images. The synthesized samples will then undergo strict manual filtering to form a reliable dataset without noticeable distortions.}
  \label{fig:dataset}
  \vspace{-0.3cm}
\end{figure}

\textbf{Step 1:} Adopt ELA \cite{li2017parallax} to stitch images from the UDIS-D dataset \cite{nie2021unsupervised} to collect extensive real stitched images. Then we dismiss those with extrapolated areas less than 10\% of the whole images.

\textbf{Step 2:} Generate abundant different mesh deformation ($w_i$) using He $et\ al.$'s algorithm \cite{he2013rectangling} from these real stitched images as shown in Fig. \ref{fig:dataset} (right).

\textbf{Step 3:} Apply the inverse of the mesh deformation ($w_i^{-1}$) to warp real rectangular images (from MS-COCO \cite{lin2014microsoft} and collected video frames) to synthetic stitched images as shown in Fig. \ref{fig:dataset} (left). The masks can be obtained by warping the all-one matrixes. Then we get triplets of real rectangular images ($R$), synthetic stitched images ($I$), and warped matrixes ($M$).


\textbf{Step 4:} Dismiss the triplets whose $I$ have distortions manually. Each manual operation will take 5-20s.
Formally, we repeat this process for three epochs and 5,705 triplets remain from more than 60,000 samples.

\textbf{Step 5:} Mix real stitched images to our training set to increase the generalization capability. Specifically, we filter out 653 samples whose $R$ has no distortion from more than 5,000 samples in step 2.


\vspace{5pt}
In sum, we prepare the DIR-D dataset with a wide range of irregular boundaries and scenes, which includes 5,839 samples for training and 519 samples for testing. Every image in the dataset has a resolution of $512\times 384$.

\section{Experiments}
\label{sec:experiments}
We first discuss experimental configuration and speed in Section \ref{subsec:Configuration}. Then we demonstrate the comparative results and ablation studies in Section \ref{subsec:Compare} and Section \ref{subsec:ablation}. 

\subsection{Experimental Configuration and Speed}
\label{subsec:Configuration}
\vspace{-5pt}
Our network is trained using an Adam optimizer \cite{kingma2014adam} with an exponentially decaying learning rate initialized to $10^{-4}$ for 100$k$ iterations. The batch size is set to 4 and we use RELU as the activation function except that the last layers of regressors adopt no activation function. $\omega_a$, $\omega_p$ and $\alpha$ are assigned as 1, $5e^{-6}$, and 0.125, respectively. $U\times V$ is set to $8\times 6$ and the implementation is based on TensorFlow. We use a single GPU with NVIDIA RTX 2080 Ti to finish all the training and inference.

It takes less than 0.4 seconds to process a 10 mega-pixel image. Similar to the experimental configuration of \cite{he2013rectangling}, the input image would be downsampled to 1 mega-pixel first and the mesh deformation is solved in the downsampled resolution. Then the mesh deformation would be upsampled and the rectangling result can be obtained by warping the full resolution input image using this upsampled deformation. The running time is dominantly on warping (interpolating) the full resolution image.

\subsection{Comparative Result}
\label{subsec:Compare}
\vspace{-5pt}
To display our superiority comprehensively, we conduct comparative experiments in quantitative comparison, qualitative comparison, user study, and cross-dataset evaluation.

\begin{table}[!t]
  \centering
  \caption{Quantitative comparisons on DIR-D.}
  \vspace{-0.3cm}
  \label{table_quanti}
  \scalebox{0.86}{
  \begin{tabular}{llll}
   \toprule
     Method & FID \cite{heusel2017gans} $\downarrow$& SSIM $\uparrow$& PSNR $\uparrow$\\
   \midrule
     Reference & 44.47 & 0.3245 & 11.30   \\
      He $et\ al.$'s. \cite{he2013rectangling} &  38.19 & 0.3775 &14.70  \\
   Ours &\textbf{21.77} &\textbf{0.7141} &\textbf{21.28}   \\
     \bottomrule
   \end{tabular}
  }
  \vspace{-0.2cm}
   \end{table}


     \begin{table}[!t]
      \centering
      \caption{No-reference blind image quality comparisons on DIR-D.}
      \vspace{-0.3cm}
      \label{blind_table}
      \scalebox{0.86}{
      \begin{tabular}{lll}
       \toprule
         Method & BIQUE \cite{venkatanath2015blind}$\downarrow$& NIQE \cite{mittal2012making}$\downarrow$ \\
       \midrule
       He $et\ al.$'s \cite{he2013rectangling} & 14.234 & 17.150  \\
       Ours & 13.989 & 16.754  \\
       Label & \textbf{11.086}& \textbf{14.872}\\
         \bottomrule
       \end{tabular}
      }
      \vspace{-0.3cm}
       \end{table}

\vspace{-12pt}
\subsubsection{Quantitative Comparison}
\vspace{-5pt}
We compare our solution with He $et\ al.$'s method \cite{he2013rectangling} on DIR-D, where 519 samples are tested for each method. We calculate the average FID\cite{heusel2017gans}, SSIM, and PSNR with labels to evaluate these solutions. The quantitative results are shown in Table \ref{table_quanti}, where `Reference' takes stitched images as rectangling results for reference.

From this table, the proposed learning solution is significantly better than the traditional solution in every metric on DIR-D. This remarkable improvement is attributed to our content-preserving property that can preserve both linear and non-linear structures.
Besides, when the location of an object in a rectangular result changes a little, it looks natural as well but the metrics might differ, which makes the quantitative experiments not completely convincing.
Therefore, we further conduct a comparison of blind image quality evaluation.
As shown in Table \ref{blind_table}, we adopt BIQUE \cite{venkatanath2015blind} and NIQE \cite{mittal2012making} as `no-reference' assessment metrics, where our solution generate higher quality results. These evaluation methods are opinion-unaware methodologies that attempt to quantify the distortion without the need for any training data. We add the evaluation of  `Label' for reference, which indicates the upper limit of the performance.


\begin{figure}[!t]
  \centering
  \includegraphics[width=0.45\textwidth]{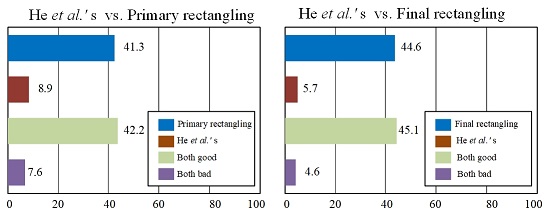}
  \vspace{-0.3cm}
  \caption{User study on DIR-D (519 testing samples). The numbers are shown in percentage and averaged on 10 participants.}
  \label{fig:user}
  \vspace{-0.3cm}
\end{figure}

\begin{figure*}
  \centering
  \begin{subfigure}{0.95\textwidth}
    \centering
    \includegraphics[width=1\textwidth, height=6cm]{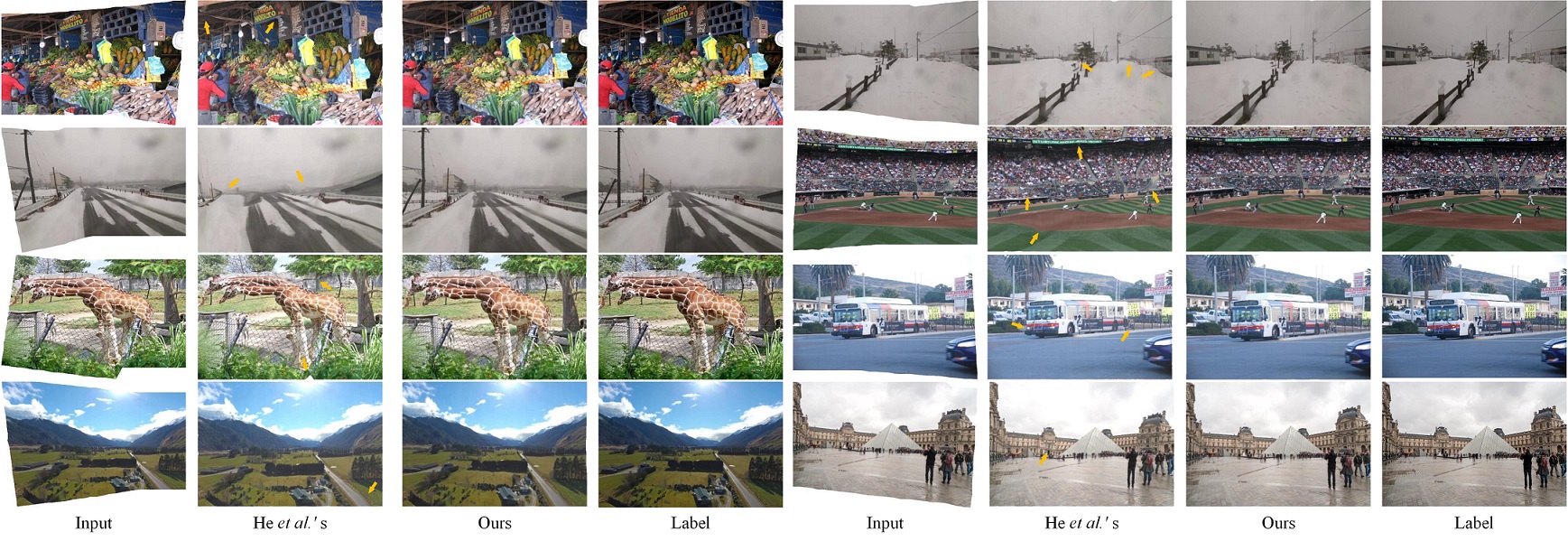}
    \caption{Scenes with linear structures.}
    \label{qua_compare_a}
  \end{subfigure}
  \hfill
  \begin{subfigure}{0.95\textwidth}
    \includegraphics[width=1\textwidth, height=6cm]{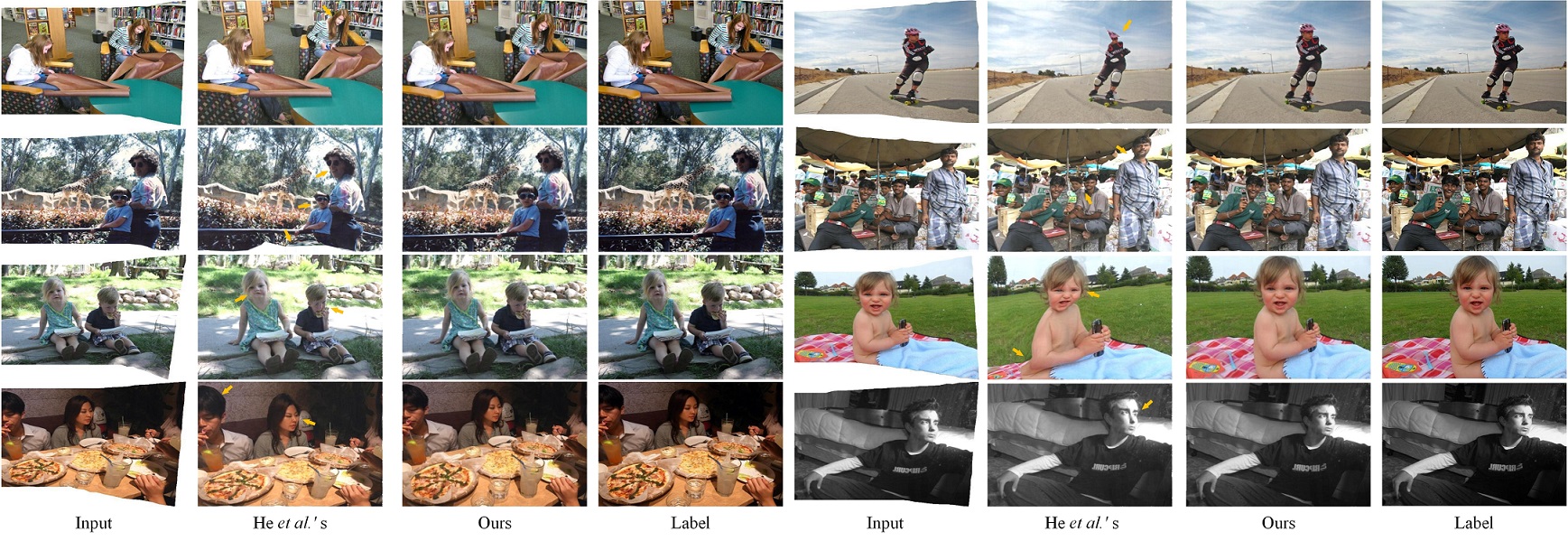}
    \caption{Scenes with non-linear structures such as portraits.}
    \label{qua_compare_b}
  \end{subfigure}
  \vspace{-0.2cm}
  \caption{Qualitative comparisons on DIR-D.}
  \label{qua_compare}
  \vspace{-0cm}
\end{figure*}


\begin{table*}[!t]
  \centering
  \caption{Ablation studies on DIR-D.}
  \vspace{-0.2cm}
  \label{table_ablation}
  \scalebox{0.95}{
  \begin{tabular}{llllllllllllll}
   \toprule
   &\multicolumn{4}{c}{Loss function}&\multicolumn{3}{c}{Mesh resolution}&\multicolumn{3}{c}{Residual progressive regression}&\multicolumn{3}{c}{Metric}\\
   \cline{2-14}
  &$\ell_c^a$  & $\ell_c^p$ & $\ell_b$& $\ell_m$& 4$\times$3 & 8$\times$6& 16$\times$12& w/o & w/ (primary)& w/ (residual)& FID $\downarrow$& SSIM $\uparrow$& PSNR $\uparrow$\\
   \midrule
     1&\checkmark  &  & \checkmark& &  & \checkmark& & \makecell[c]{\checkmark}& & & 115.37& 0.3498& 14.43\\
     2&\checkmark  & \checkmark & \checkmark& &  & \checkmark& & \makecell[c]{\checkmark}& & & 24.57& 0.6109& 19.84\\
     3&\checkmark  & \checkmark & \checkmark& \checkmark&  & \checkmark& & \makecell[c]{\checkmark}& & & 22.43& 0.6926& 20.92\\
     4&\checkmark  & \checkmark & \checkmark& \checkmark& \checkmark & & & \makecell[c]{\checkmark}& & & 24.15& 0.6361& 20.16\\
     5&\checkmark  & \checkmark & \checkmark& \checkmark&  & &\checkmark & \makecell[c]{\checkmark}& & & 22.32& 0.6907& 20.95\\
     6&\checkmark  & \checkmark & \checkmark& \checkmark&  & &\checkmark & & \makecell[c]{\checkmark}& & 22.35& 0.6902& 20.93\\
     7&\checkmark  & \checkmark & \checkmark& \checkmark&  & &\checkmark & & \makecell[c]{\checkmark}& \makecell[c]{\checkmark}& \textbf{21.77}& \textbf{0.7141}& \textbf{21.28}\\
      \bottomrule
   \end{tabular}
  }
   \vspace{-0.1cm}

   \end{table*}

\vspace{-10pt}
\subsubsection{Qualitative Comparison}
\vspace{-5pt}
\label{quali_1}
To compare the qualitative results comprehensively, we divide the testing set into two parts. The first part includes abundant linear structures that are suited to the traditional baseline, while the second one includes extensive non-linear structures such as portraits.

From the results in Fig. \ref{qua_compare}, we can observe that our method significantly outperforms the traditional solution in the two scenes. We owe our superiority to the content preserving capability that can keep the mesh shape-preserving and content perceptually natural. The traditional solution does not perform well in scenes with linear structures due to the limited line detection capability. The failure occurs in portraits with non-linear objects because non-linear properties are not included in its optimized energy.


\vspace{-10pt}
\subsubsection{User Study}
\vspace{-5pt}
\label{user_study}
The motivation of image rectangling is that the users are not satisfied with the irregular boundaries in stitched images. Therefore, our goal is to produce rectangular images that please the most users.

We conduct user studies on visual preference. Formally, we compare He $et\ al.$'s algorithm with our primary rectangling and final rectangling (as shown in Fig. \ref{fig:network}) one by one.
At each time, three images are shown on one screen: the input, He $et\ al.$'s rectangling, and ours (primary or final). We shuffle the order of different methods each time. The users may zoom in on the images and are required to answer which result is preferred.
In this study, we invite 10 participants, including five researchers/students with computer vision backgrounds and five volunteers outside this community. The results are shown in Fig. \ref{fig:user}, where our solution is preferred by more users.

\vspace{-10pt}
\subsubsection{Cross-Dataset Evaluation}
\vspace{-5pt}
\label{quali_2}
In this cross-dataset evaluation, we adopt the DIR-D dataset to train our model and test this model in other datasets. 

\begin{figure*}
  \centering
  \begin{subfigure}{0.95\textwidth}
    \centering
    \includegraphics[width=1\textwidth, height=12cm]{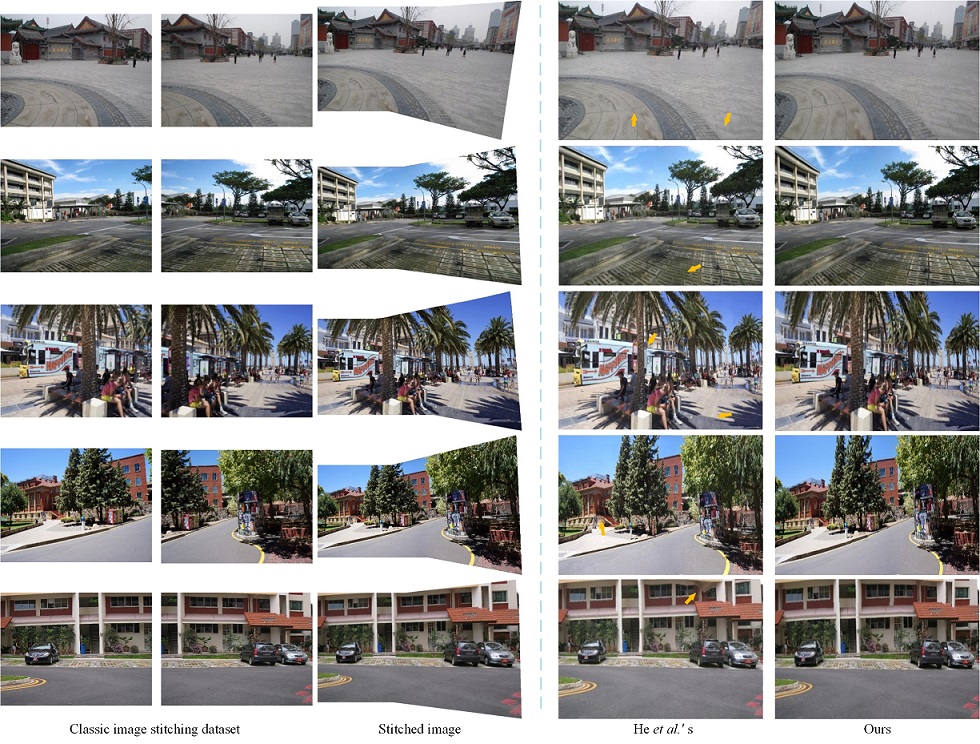}
    \caption{Rectangling high-quality stitched images. The stitching datasets are from \cite{gao2011constructing, zaragoza2013projective,gao2013seam}, and the stitching algorithms are from \cite{nie2021unsupervised,jia2021leveraging}.}
    \label{cross_dataset_a}
  \end{subfigure}
  \hfill
  \begin{subfigure}{0.95\textwidth}
    \includegraphics[width=1\textwidth, height=2.7cm]{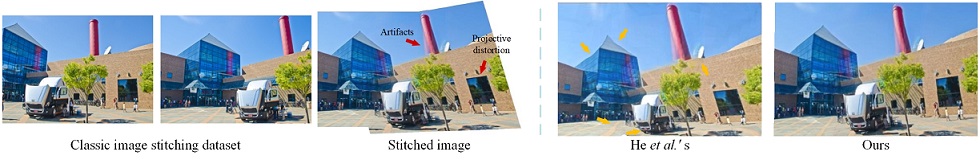}
    \caption{Rectangling low-quality stitched images, in which artifacts and distortions can be found. We adopt \cite{nie2021unsupervised} to stitch images from \cite{zhang2014parallax}. We discuss this low-quality stitching example to show the effectiveness of our method in general real stitching scenes.}
    \label{cross_dataset_b}
  \end{subfigure}
  \vspace{-0.2cm}
  \caption{Cross-dataset qualitative comparisons. The arrows highlight the distorted regions.}
  \label{cross_dataset}
  \vspace{-0.2cm}
\end{figure*}

Formally, we adopt different image stitching methods (SPW \cite{liao2019single}, LCP \cite{jia2021leveraging}, and UDIS \cite{nie2021unsupervised}) to stitch classic image stitching datasets \cite{gao2011constructing, zaragoza2013projective, jia2021leveraging, gao2013seam}. Then, stitched images are used for rectangling using different algorithms. The results are shown in Fig. \ref{cross_dataset_a}, where our solution produces fewer distortions in rectangling results.

To show our effectiveness in more general scenes, where artifacts and projective distortions are not eliminated, we demonstrate rectangling results on a failure case of image stitching. As shown in Fig. \ref{cross_dataset_b}, our method still works well.

\begin{figure}[!t]
  \centering
  \includegraphics[width=0.47\textwidth, height=2.5cm]{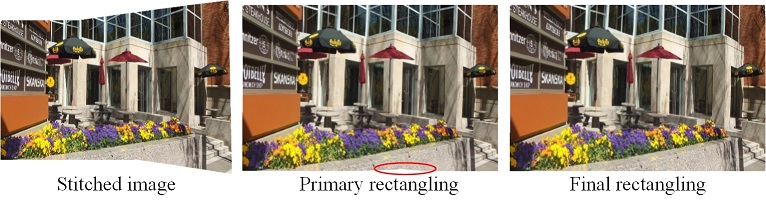}
  \vspace{-0.3cm}
  \caption{Ablation study of the residual progressive regression strategy on other dataset (`potberry'\cite{jia2021leveraging}). The circles highlight the regions with uneven boundaries.}
  \label{fig:ablation}
  \vspace{-0.3cm}
\end{figure}

\subsection{Ablation Studies}
\label{subsec:ablation}
\vspace{-0.3cm}

  The proposed network is simple but effective. We validate the effectiveness of every module on DIR-D.

  \textbf{Loss function.} We ablate the residual regressor as the baseline structure and evaluate the effectiveness of different constraint terms in our objective function. As shown in experiment 1-3 of Table \ref{table_ablation}, both the content term and mesh term can significantly improve our performance.

  \textbf{Mesh resolution.} We test different mesh resolutions of $4\times 3$, $8\times 6$ and $16\times 12$. As shown in the experiment 3-5 of Table \ref{table_ablation}, $4\times 3$ mesh decreases the rectangling performance while $8\times 6$ mesh and $16\times 12$ mesh give similar results. Nevertheless, $16\times 12$ mesh brings more computational costs, thus we adopt the $8\times 6$ mesh in our implementation.

  \textbf{Residual progressive regression.} We validate the effectiveness of our residual progressive regression strategy in experiment 6-7 of Table \ref{table_ablation}, where the residual regressor continues to refine the rectangling results based on the primary regressor. Although the improvement on the DIR-D dataset is slight, this strategy enhances our generalization capability to avoid the uneven boundaries in other datasets as shown in Fig. \ref{fig:ablation}.

\vspace{-4pt}
\section{Conclusion}
\label{sec:conclusion}
\vspace{-4pt}
In this paper, we propose the first deep image rectangling solution and dataset for image stitching. Compared with traditional two-stage methods, the proposed solution is a one-stage method, enabling efficient parallel computation with a predefined rigid target mesh. Besides, our solution can preserve both linear and non-linear structures, demonstrating the superiority over the existing methods both quantitatively and qualitatively.


{\small
\bibliographystyle{ieee_fullname}
\bibliography{egbib}
}

\newpage
\appendix




\begin{figure}
  \centering
  \begin{subfigure}{0.5\textwidth}
    \centering
    \includegraphics[width=0.98\textwidth]{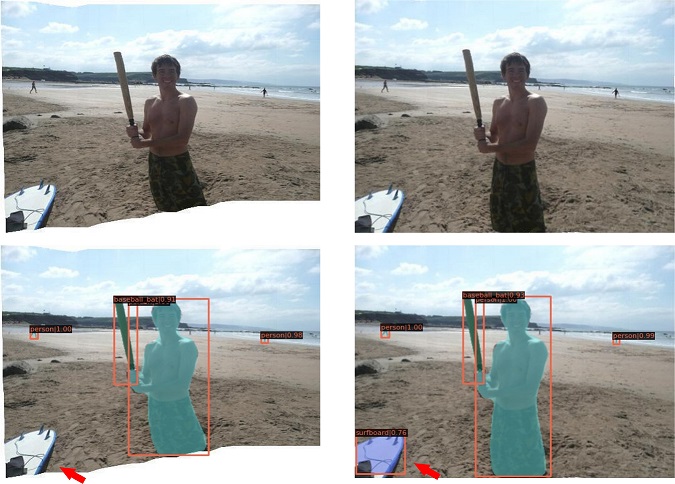}
    \caption{Missing `surfboard’.}
    \label{seg-a}
  \end{subfigure}
  \hfill
  \centering
  \begin{subfigure}{0.5\textwidth}
    \centering
    \includegraphics[width=0.98\textwidth]{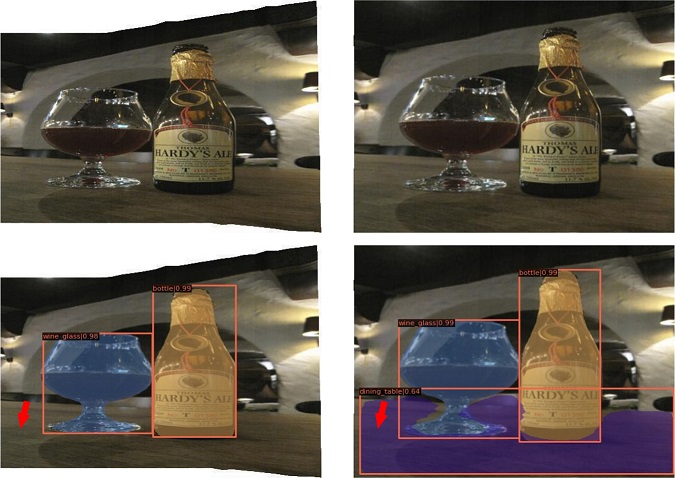}
    \caption{Missing `dining table'}
    \label{seg-b}
  \end{subfigure}
  \hfill
  \centering
  \begin{subfigure}{0.5\textwidth}
    \centering
    \includegraphics[width=0.98\textwidth]{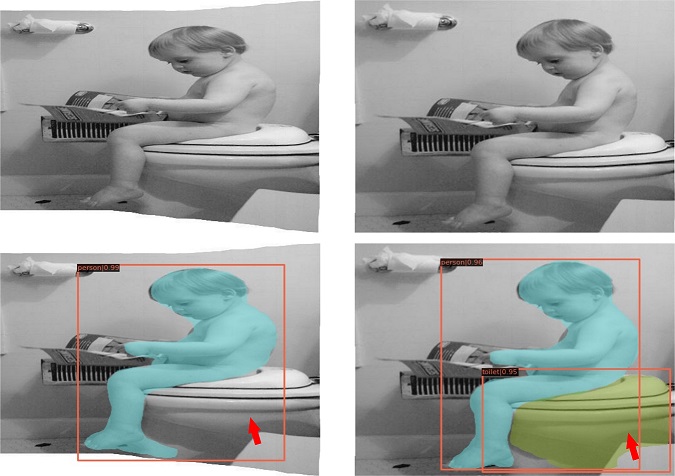}
    \caption{Missing `toilet'}
    \label{seg-b}
  \end{subfigure}
  \vspace{-0.2cm}
  \caption{Object detection and semantic segmentation results of the stitched images and our rectangling results. The arrows highlight the missing parts.}
  \label{segmentation}
  \vspace{-0.3cm}
\end{figure}

\begin{figure*}[!t]
  \centering
  \includegraphics[width=0.97\textwidth]{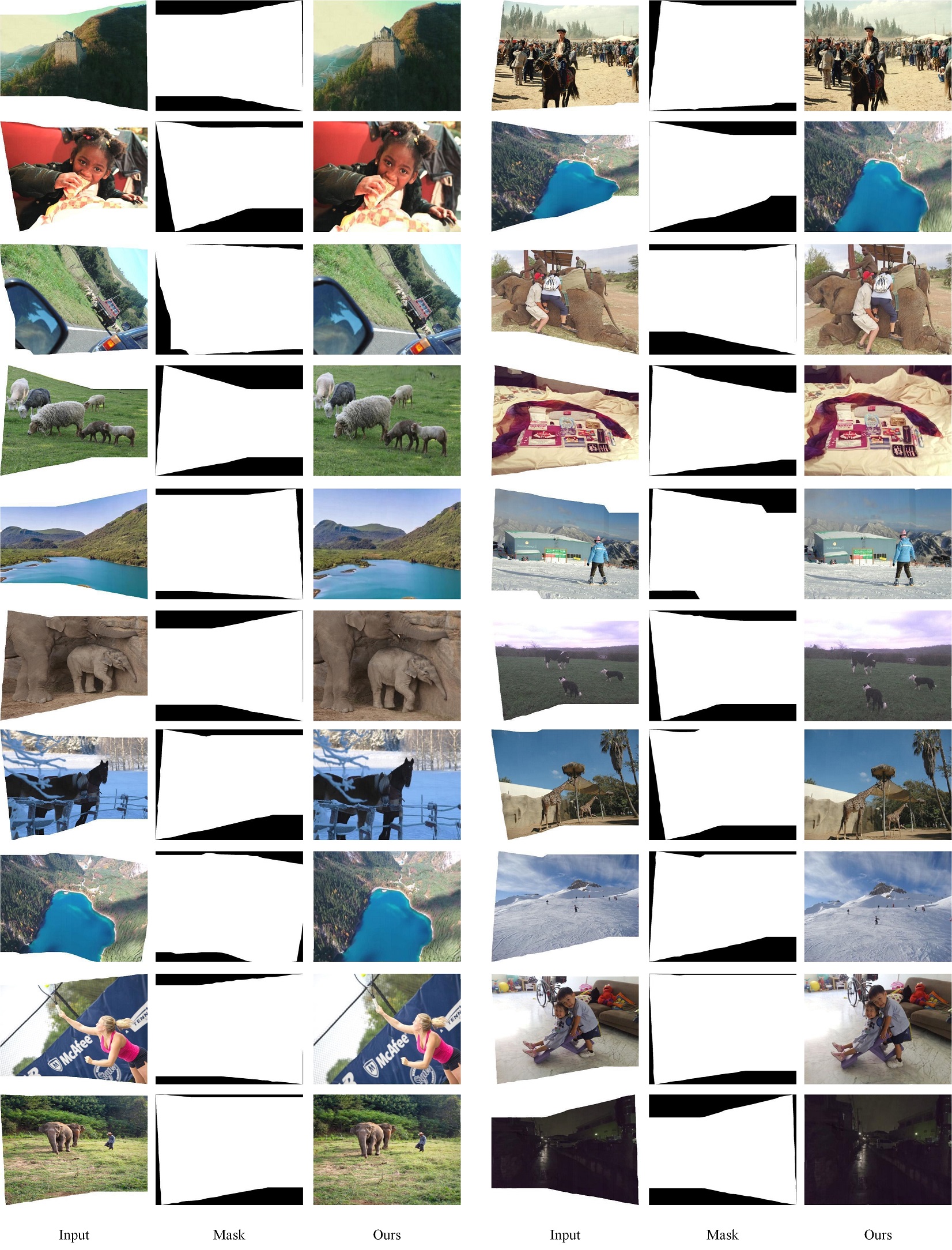}
  \caption{More results of our solution on DIR-D. Each triplet includes an input, a mask, and our rectangling result from left to right.}
  \label{fig:fig1}
\end{figure*}

\begin{figure*}[!t]
  \centering
  \includegraphics[width=0.92\textwidth, height=21.5cm]{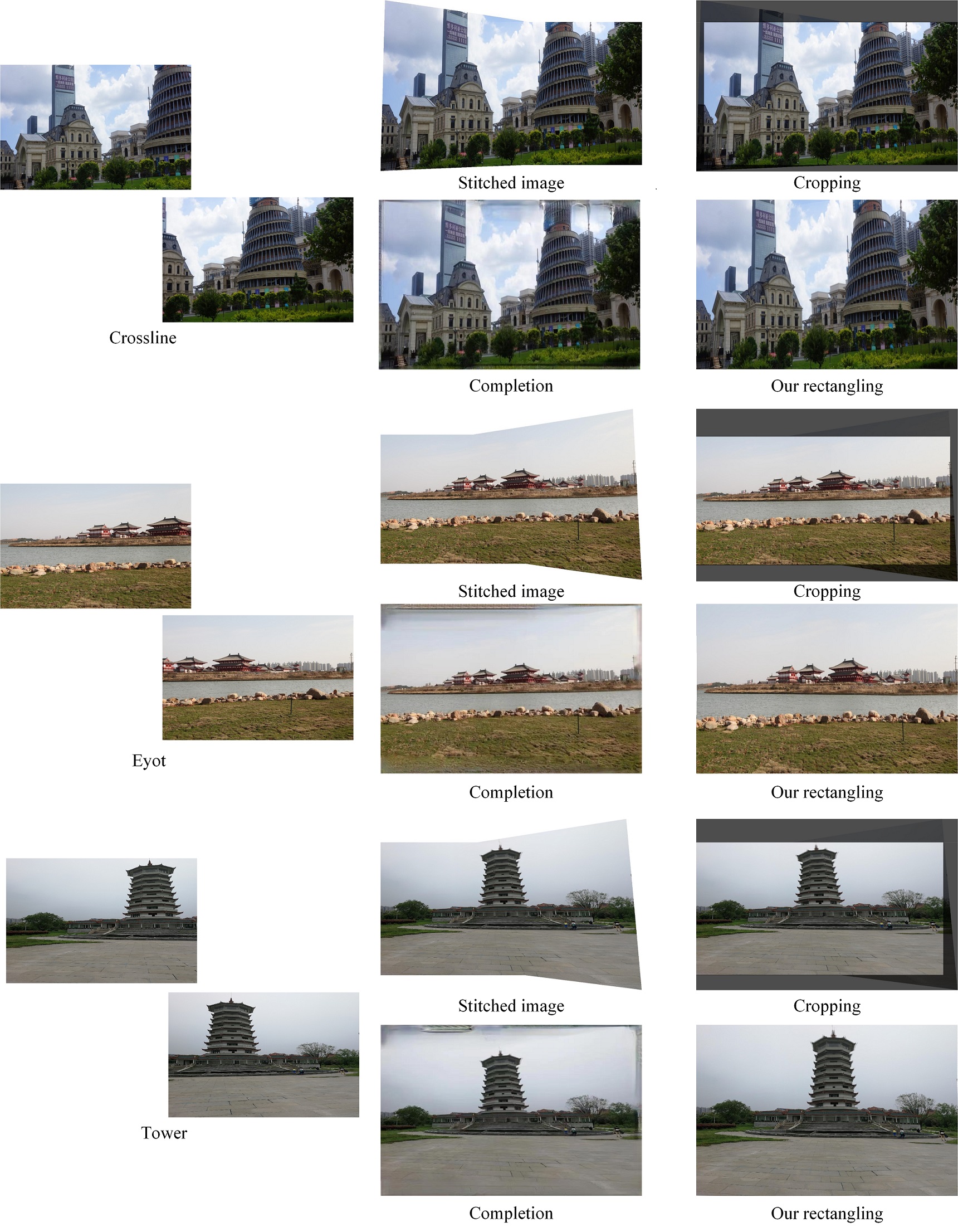}
  \caption{More cross-dataset results. The classic image stitching datasets (`crossline' \cite{jia2021leveraging}, `eyot' \cite{jia2021leveraging} and `tower' \cite{li2017parallax}) are stitched by SPW \cite{liao2019single}. We adopt LaMa \cite{suvorov2021resolution} to complete the stitched images, and the rectangling results are generated by the proposed learning baseline.}
  \label{fig:fig2}
\end{figure*}

\section{Overview}
\label{sec:Overview}

In the supplementary material, we first demonstrate the benefits of image rectangling for scene reasoning in Section \ref{sec:Benefits}. Then, we illustrate more experimental results of our solution in Section \ref{sec:Experiment}, including our rectangling results on the DIR-D dataset and other datasets.

\section{Benefits for Scene Reasoning}
\label{sec:Benefits}

The proposed rectangling solution offers a nearly perfect visual perception for users by eliminating the irregular boundaries in image stitching. It can also help downstream vision tasks such as object
detection and semantic segmentation, which is crucial for scene understanding. As shown in Fig. \ref{segmentation}, the detection and segmentation results are derived from Mask R-CNN \cite{8237584}. We can notice that the objects in the stitched images with irregular boundaries may be missing, such as the surfboard (Fig. \ref{segmentation}{\color{red}a}), the dining table(Fig. \ref{segmentation}{\color{red}b}), and the toilet (Fig. \ref{segmentation}{\color{red}c}). By contrast, our rectangling results `find' the missing objects. We summarize the improvement as follows:



Almost all existing deep learning models (detection and segmentation) are trained on rectangular images, making them not robust to the regions around the boundaries in stitched images.

\section{More Results}
\label{sec:Experiment}

More results on DIR-D are exhibited in Fig. \ref{fig:fig1}, where our solution can deal with variable irregular boundaries and yield perceptually natural rectangular results.

Besides, more cross-dataset results are displayed in Fig. \ref{fig:fig2}, which shows the superiority of rectangling over other solutions such as cropping and completion.


\end{document}